\documentclass{article}

\usepackage[final]{nips_2017}

\usepackage[utf8]{inputenc} 
\usepackage[T1]{fontenc}    
\usepackage{hyperref}       
\usepackage{url}            
\usepackage{booktabs}       
\usepackage{amsfonts}       
\usepackage{nicefrac}       
\usepackage{microtype}      
\usepackage{graphicx}
\usepackage{tabularx}
\usepackage{comment}
\usepackage{times}
\usepackage{alltt}
\usepackage{color}
\usepackage{enumitem}
\usepackage{upquote,textcomp}
\usepackage{fancyvrb}

\usepackage{amsbsy,xspace}
\usepackage{booktabs,longtable,rotating} 
\usepackage{array}
\usepackage{multirow} 
\usepackage{fancyhdr} 
\usepackage{setspace} 
\usepackage{url} 
\usepackage{microtype} 

\usepackage{type1cm}
\usepackage{lettrine}
\usepackage{backref} 
\usepackage{emptypage}

\def\cpt{CPT\xspace}
\def\icd{ICD\xspace}
\def\cpts{CPTs\xspace}
\def\icds{ICDs\xspace}

\title{Intelligent EHRs: Predicting Procedure Codes From Diagnosis Codes}

\author{Hasham Ul Haq\\
  MTBC\\
  \texttt{hashamulhaq@mtbc.com}\\
  \And
  Rameel Ahmad\\
  MTBC\\
  \texttt{rameelahmad@mtbc.com}\\
  \And
  Sibt Ul Hussain\\
  Reveal Lab, NUCES-FAST\\
  \texttt{sibtul.hussain@nu.edu.pk}
}  
\begin{document}
\maketitle

\begin{abstract}
 In order to submit a claim to insurance companies, a doctor needs to code a patient encounter with both the diagnosis (\icds) and procedures  performed (\cpts) in an Electronic Health Record (EHR). Identifying and applying relevant procedures code is a cumbersome and time-consuming task as a doctor has to choose from around 13,000 procedure codes with no predefined one-to-one mapping.  In this paper, we propose a state-of-the-art deep learning method for automatic and intelligent coding of procedures (\cpts) from the diagnosis codes (\icds) entered by the doctor. Precisely, we cast the learning problem as a multi-label classification problem and use distributed representation to learn the input mapping of high-dimensional sparse \icds codes. Our final model trained on 2.3 million claims is able to outperform existing rule-based probabilistic and  association-rule mining based methods and has a recall of 90@3. 
 

\end{abstract}

\section{Introduction}
In order to be reimbursed for their services provided to patients,  doctors (providers) need to provide proof of the procedures that they performed. To do this each patient visit needs to be coded with appropriate diagnosis and procedure codes before submitting a claim to the insurance companies. Assigning correct procedure codes is vital because providers are paid and evaluated according to each procedure they perform on the patient. However, assigning correct procedure codes is a challenging tasks and involve complete attention to details from the doctors. This coding task is further complicated by the fact that: (i) some insurance companies (like Medicare) have their own coding rules and only accept claims containing certain codes; (ii) some insurance companies do not accept a certain combination of diagnosis and procedure codes; (iii)  doctors can be penalized for over-coding (i.e. assigning a code for a more serious condition than it is justified thus causing financial loss to insurance companies); (iv) doctors can incur financial loss for under-coding (under-coding happens by missing out pertinent codes for procedures that they have performed); (iv) different age and gender necessitates different set of procedural codes; (v) some codes depend on the duration of encounter not on the prognosis of a disease; and finally (vi) applying wrong code can result in claim denial. 

Thus providers need to be extremely careful and add in all the factors while coding for procedure codes. This not only consumes a lot of their valuable time but lead to loss of their focus on patient. Unfortunately there exists no such method, according to our best of knowledge, that can reliably and intelligently predict the relevant procedure codes from the given diagnosis, age, gender, etc.

To this end, we propose a state-of-the-art deep learning based method for predicting the \cpts from \icds. Precisely, we make following two main contributions. Firstly, we introduce a distributed-representation \citep{word2vec} for learning the dense-embeddings from variable number of sparse \icds. Secondly, we cast the problem of \cpts prediction as multi-label classification and use sigmoid-loss function in a deeply connected network to learn the \icds to \cpts mappings.

Our proposed model is trained on a large scale dataset (with 2.3 Million) and gives state of the art performance relative to probabilistic and association-rule mining based methods. This model has already been integrated into an EHR and leads to significant reduction in doctors time during the application of relevant \cpt codes.

We discuss our model in section \ref{model} while experiments and data set details are given in section \ref{experiments}. Section \ref{results} contains results and discussion. We conclude the paper with related work in \ref{rwork}

\section{Neural Network Architecture}\label{model}

In this section we introduce our Neural Network architecture. The biggest challenge we faced while designing our network architecture was due to the categorical nature of our input data. For example, total number of diagnosis codes are $\approx 70,000$. One of the simplest method of dealing with categorical data is to encode it using one-hot
encoding, but when the number of categories is too large, one-hot encoding no longer remains feasible as the resulting matrix becomes too sparse and with limited continuous features, training a neural network on a large 
categorical data set results in limited performance. For this problem we had to find a new method of encoding categorical data.

The structure of ICD-10 codes provide a unique solution to this problem. As ICD-10 codes are limited to 7 characters -- where each character can either be an alphabet or an integer i.e 36 unique values -- we can encode each character of the ICD code independently to generate a less sparse representation for each diagnosis. Precisely, our model learns a separate embedding matrix for each character of the input \icd. These embeddings are then concatenated to get a dense representation of complete \icd. For a variable number of \icds, our models generate dense embeddings for each \icd and then these dense embeddings are added to form a single representation. This representation is then given input to next affine layer in the model. Overall, our model has one embedding and 4 densely-connected affine layers.

\begin{figure}
  \includegraphics[scale=0.5]{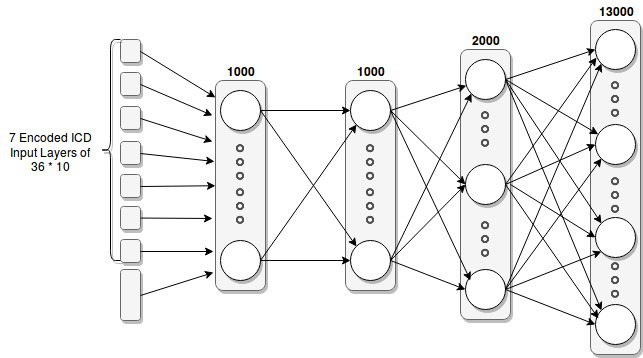}
  \caption{Neural Network Architecture}
  \label{fig:NetworkArchitecture}
\end{figure}

In our data exploration phase, we had discovered that each provider has his/her own behavior of applying procedures and thus initially decided to train separate models for each provider based on his historical data. Since in our data we have $\approx 2100$ unique providers, it becomes practically infeasible to build and deploy such large number of models. To circumvent this in our this neural network, we added an additional embedding layer to incorporate the behavior of each provider. Input to this layer was one-hot encoded matrix for each provider identification number. With the introduction of this layer the neural network was able to distinguish between different providers which enabled us to achieve much higher accuracy. To encapsulate the demands and behaviour of insurance companies we added an additional embedding layer. Input to this layer was one-hot encoded matrix for each insurance id. However, this technique produced neglectable performance gain, so we eliminated it from our final architecture.
Figure \ref{fig:NetworkArchitecture} explains the architecture of our Neural Network.

The first layer consists of input layers for \icd and provider codes. The input layers are then stacked vertically and fed to a network of 4 fully connected layers. The final layer has output dimension equal to the number of classes.  We use ReLU nonlinearity after each affine transformation. In the testing phase we also apply sigmoid nonlinearity after the final layer to get probabilities for each class. 

Finding the best loss-function for multi-labeling also proved to be quiet a challenging task. We tested different loss-functions like softmax cross-entropy, hamming loss and sigmoid base cross entropy loss. However, extensive testing showed sigmoid based cross-entropy loss to be the best loss-function for our problem at hand. Specifically, for a given claim, we attach sigmoid at the output of each neuron and then take the negative of sum of log-likelihood of all the true \cpts as our loss. We average this loss across all the batch examples to obtain final loss function.

As a post-processing we apply fixed age and gender rules on the predicted \cpts to filter out any miss-predictions.

\section{Experiments}\label{experiments}
\subsection{Data}
Our data set consisted of 2.3 Million claims submitted to different insurance companies from a U.S-based billing company. The claim data comprised of patient and provider's information, diagnosis (ICD-10) and procedure codes (\cpts). We used only ICD-10 codes and only those claims were used that had already been accepted and paid by insurance companies, this ensures that our training data has minimum possible noise and contains only authentic claims. Also, this gave us the opportunity to observe and verify rules set up by different insurance companies like Medicare.

Upon observing the data we discovered that the procedure codes contained codes that are not dependent on diagnosis codes like Physician Quality Reporting System  (PQRS) codes. We also discovered that visit and vaccination codes are recorded by following a standard procedure so we decided to build separate rule-based models to automate them.
\subsection{Probabilistic Prediction Model With Rule based Pruning}
In addition to our neural network, in our initial iterations for building an automated method we also tested a probabilistic model. This probabilistic model was built on the Bayes rule. In this model we tried to predict the probability of a \cpt given \icd and other relevant information, i.e. $P(\cpt | \icd, gender, age)$. The likelihood and prior probabilities were calculated from historical records. This model was giving good performance however we found out that it was extremely dependent on the availability of provider and specialization data and it could not generalize well to new practices. Furthermore, getting multiple \cpts predictions from given \icds required too much manual tuning, crafting and domain knowledge. 

\subsection{Association Rule Mining}
In our initial version of the solution, we also mapped the \cpt prediction problem as frequent-item set mining. For this purpose we used the apriori \citep{pasquier1999efficient} algorithm. In this case an item consisted of diagnosis and procedure codes in each claim. 

After generating frequent item sets we generated rules along with their confidence by normalizing the counts of unique diagnosis items and their corresponding set of procedure codes. During the testing phase we looked for highest matching set of diagnosis codes and finally procedure codes with highest confidence was selected. The predicted codes were then filtered through age and gender rules.

Using association rule mining on such a large data set proved to be quite challenging. As searching for rules on large data set required a lot of time, memory, and computational power. 
\section{Results \& Conclusions} \label{results}
For our training data we used 2.3 Million patient claims submitted to different insurance companies, while for test set we used $\approx 70,000$ claims of different providers submitted to different insurance companies. For training we used Adam optimizer with adaptive learning rate. 

To evaluate the performance of our system, we have used precision-recall as our evaluation metric. Precisely we report recall and precision @ K.  Table ~\ref{tab:res} compares the results of our proposed neural network method with probabilistic and rule-mining methods. 

Our proposed deep learning model is able to achieve the best precision and recall. The biggest advantage of this method is that it remains generic and can be trained end-to-end to obtain multi-label prediction of \icds. In comparison our probabilistic model, while remaining fast to train, achieves low precision and recall values and is relatively harder to tune and is not too generic. 

Overall, we would recommend our deep learning model for predicting \cpts from \icds, which is not only easy to train but is more generalizable and remains the best performer in our large scale experiments.
%
\begin{table}[t]
  \caption{Comparative results of different methods built. Our deep learning method is giving the best performance while being generic and an end-to-end model.}
  \label{tab:res}
  \centering
   \resizebox{\linewidth}{!}{
  \begin{tabular}{llllll}
    \toprule
    Name     & Training Time     & Recall @ 3 & Precision @ 3 \\
    \midrule
    Probabilistic model + Rules  & 20 minutes  & 85   & 37\\
    Deep Learning Model    & 5 hours  & \textbf{90} & \textbf{45}  \\
    Apriori   & 48 hours & 70  & 20\\
    \bottomrule
  \end{tabular}}
\end{table}

\section{Related Work}\label{rwork}
Multi-label Classification has been a standard problem in machine learning. However, it has rarely been employed for the automatic assignment of procedure codes.  Closely related work to ours is the automation of diagnosis process and future health status prediction of patients using LSTMs (Long Short Term Memory), RNNs (Recurrent Neural Networks),\cite{lecun2015deep} and Convolutional Neural Networks \cite{lipton2015learning}. Although, these papers introduce new techniques for dealing with the high dimensionality of clinical data like auto encoders and embedding learning \cite{kuchaiev2017training}, they mostly focus on the historical record based modeling of a single existing patient. As far as predicting \icds from \cpts, according to our best of knowledge no such work exist which has tackled this problem before.

\bibliographystyle{apalike}
\bibliography{biblio,deep-learning}

\begin{thebibliography}{}

\bibitem[Kuchaiev and Ginsburg, 2017]{kuchaiev2017training}
Kuchaiev, O. and Ginsburg, B. (2017).
\newblock Training deep autoencoders for collaborative filtering.
\newblock {\em arXiv preprint arXiv:1708.01715}.

\bibitem[LeCun et~al., 2015]{lecun2015deep}
LeCun, Y., Bengio, Y., and Hinton, G. (2015).
\newblock Deep learning.
\newblock {\em Nature}, 521(7553):436--444.

\bibitem[Lipton et~al., 2015]{lipton2015learning}
Lipton, Z.~C., Kale, D.~C., Elkan, C., and Wetzell, R. (2015).
\newblock Learning to diagnose with lstm recurrent neural networks.
\newblock {\em arXiv preprint arXiv:1511.03677}.

\bibitem[Mikolov et~al., 2013]{word2vec}
Mikolov, T., Chen, K., Corrado, G., and Dean, J. (2013).
\newblock Efficient estimation of word representations in vector space.
\newblock {\em arXiv preprint arXiv:1301.3781}.

\bibitem[Pasquier et~al., 1999]{pasquier1999efficient}
Pasquier, N., Bastide, Y., Taouil, R., and Lakhal, L. (1999).
\newblock Efficient mining of association rules using closed itemset lattices.
\newblock {\em Information systems}, 24(1):25--46.

\end{thebibliography}

\end{document}